\documentclass[conference]{IEEEtran}
\IEEEoverridecommandlockouts

\usepackage{boldline} 

\usepackage{amsmath,amsfonts}

\usepackage[english]{babel}
\usepackage[
    backend=biber,
    style=ieee,
    autocite=inline,
    maxcitenames=2,
    mincitenames=1,
]{biblatex}
\usepackage{csquotes}

\bibliography{ref.bib}
\DefineBibliographyStrings{english}{%
  bibliography = {References},}
\usepackage{blindtext}
\usepackage{pdfpages}

\usepackage{hyperref}
\usepackage{enumitem}
\usepackage{multirow}
\usepackage{tabularx}

\usepackage{amsmath,amssymb,amsfonts}
\usepackage{algorithmic}
\usepackage{subfig}
\usepackage{textcomp}
\usepackage{xcolor}
\usepackage{array,graphics,graphicx,ifthen,keyval,makecell,multirow,rotating,trig}
\usepackage{graphicx}

\def\BibTeX{{\rm B\kern-.05em{\sc i\kern-.025em b}\kern-.08em
    T\kern-.1667em\lower.7ex\hbox{E}\kern-.125emX}}

\begin{document}

\title{Domain Adaptation for Facial Expression Classifier via Domain Discrimination and Gradient Reversal}

\author{
    \IEEEauthorblockN{Kamil Akhmetov}
    \IEEEauthorblockA{
        \textit{Institute of Data Science \& AI} \\
        \textit{Innopolis University}\\
        Innopolis, Russia \\
        k.ahmetov@innopolis.ru
    }
}

\maketitle

\begin{abstract}

Bringing empathy to a computerized system could significantly improve the quality of human-computer communications, as soon as machines would be able to understand customer intentions and better serve their needs. According to different studies (\nameref{sec:lr}),  visual information is one of the most important channels of human interaction and contains significant behavioral signals, that may be captured from facial expressions. Therefore, it is consistent and natural that the research in the field of Facial Expression Recognition (FER) has acquired increased interest over the past decade due to having diverse application area including health-care, sociology, psychology, driver-safety, virtual reality, cognitive sciences, security, entertainment, marketing, etc. We propose a new architecture for the task of FER and examine the impact of domain discrimination loss regularization on the learning process. With regard to observations, including both classical training conditions and unsupervised domain adaptation scenarios, important aspects of the considered domain adaptation approach integration are traced. The results may serve as a foundation for the further research in the field.

\end{abstract}


\section{Introduction} \label{sec:intro}
Countless amount of factors (pose, brightness, background, occlusions, subject ethnicity, face shape, etc.) make Facial Expression Recognition (FER) a complex Computer Vision (CV) problem \cite{Li2018DeepFE}. 
Conventionally, manual Face Detection (FD) and handcrafted Feature Extraction (FE) designed by domain experts involved traditional Machine Learning (ML) techniques for emotion classification \cite{Li2018DeepFE, Corneanu2016SurveyOR, Wu2018FacialLD}. Later, automatic FER has moved to Deep Learning (DL), which has emerged as a general approach for ML tasks. Convolutional Neural Networks (CNN) followed by Fully Connected (FC) classification layers reduce the need for pre-processing and enable ‘end-to-end’  learning. 

Modern research focuses on domain agnostic models developed for real-world scenarios. Domain Adaptation approaches mainly focus on the segregation between domain-specific and task-specific (domain-independent) features from the original data. However, according to \citeauthor{Siddiqi2016EvaluatingRP} modern approaches are still weak to face real-world scenarios \cite{Siddiqi2016EvaluatingRP}. Thus, further studies focus on discovering new ways of learning facial expressions improving performance and required computation cost.

The paper summarizes the key findings of the past research, introduces a new architecture for FER using transfer learning, and examines the applicability of domain discrimination loss.
The rest of the work is organized per the following sections: Literature Review, Methodology, Evaluation \& Discussion and Conclusion.

\section{Literature Review} \label{sec:lr}

The section reviews the most relevant literature in the appropriate subfields and finishes with a summary of the key findings, identifies the research gap, and introduces the research questions.

\subsection{Facial Expression Recognition} \label{subsec:fer}
Historically, works in the field of FER are grouped as follows: Hand Crafted Features (HCFs) based models (also known as Classical Machine Learning) and Deep Learning (DL) approaches. With the emergence of deep learning, traditional methods became comparatively less used due to limitations covered by CNN based models, that generally provide competitive (with state-of-the-art) results and allow learning shared representations, which is especially important for Domain Adaptation.

\begin{itemize}
    \item 
        \textbf{Classical FER} \\ 
        Hand-Crafted Features and traditional classifiers are less expensive in terms of resources, while having less expressiveness power. Learned patterns are easier to analyze and understand causal relationships. \cite{Schapire2013explaining, Liaw2002RandomForest, Cortes1995Support, Chen2014FacialER, Shan2009FacialER, Bartlett2005RecognizingFE, Lyons2020CodingFE, Whitehill2006HaarFF, Khan2013FrameworkFR, Ghimire2016FacialER, Suk2014RealTimeMF, Ghimire2013GeometricFF, BenitezQuiroz2016EmotioNetAA, Torre2015IntraFace, Siddiqi2013HierarchicalRS, Siddiqi2014HumanFE, Siddiqi2014DepthCF, Siddiqi2014FacialER, Siddiqi2015HumanFE, Siddiqi2016ANM}
    \item 
        \textbf{Attentional Deep Learning based FER} \\
        Focus attention during learning on particular regions since only specific details contribute to each of the facial expressions. Hence, give more weight to the features containing more information (statistically significant). 
        \cite{Khorrami2015DoDN, Susskind2010toronto, Memisevic2015ZerobiasAA, Paine2015AnAO, Minaee2019DeepEmotionFE, Lucey2010ExtendedCK, Lyons2020CodingFE, Li2019OcclusionAF}
    \item
        \textbf{Disentangled representation learning} \\
        Disentangle data into task significant features while reducing the impact of non-important sources of variation. The caused separation facilitates domain generalization via moving feature space to higher levels of abstraction. 
        \cite{Bengio2014representation, Hinton1993AutoencodersMD, Liu2018ExploringDF, Halawa2020learning, Li2019OcclusionAF, Halawa2020learning}
    \item
        \textbf{Identity-Aware CNN} \\
        Utilize the knowledge on identities to remove the bias introduced by personal characteristics of a subject. Separating identity impact yields clearer expression related features, thus improving the learning generalization.
        \cite{Liu2018ExploringDF, Simonyan2015Deep, Tu2019IdenNetIF, Mavadati2013DISFA, Zhang2018IdentitybasedAT}
    \item
        \textbf{Generative Adversarial Networks} \\ 
        Adversarial learning nature applied for the task of FER yields models having higher domain adaptation, being though more complex. Features learned in adversarial process may be simultaneously exploited for expression classification.
        \cite{Cai2019identityfree, Ali2019facial, Lucey2010ExtendedCK, Arjovsky2017WassersteinGA,  Xu2019PersonindependentFE, Kim2017DeepGN, Makhzani2016adversarial, Susskind2010toronto}
    \item
        \textbf{Other single frame based methods} 
        Approaches include suggest sophisticated \textit{data sources} (stylized character samples and utilizing non-peak expressions), model architectures (inception layers) and learning process design (teacher and student instances).
        \cite{Aneja2016ModelingSC, Lucey2010ExtendedCK, Lundqvist2015KarolinskaDE, Pantic2005wdffe, Mollahosseini2016GoingDI, Gross2008MultiPIE, , Mavadati2013DISFA, Goodfellow2015ChallengesIR, Liu2014FacialER, Lyons2020CodingFE, Zhao2016PeakPilotedDN, Szegedy2015GoingDW, Ding2017FaceNet2ExpNetRA, Ba2014DoDN, Parkhi2015DeepFR, Cugu2019MicroExpNetAE, Li2013OuluCasia, Siqueira2020EfficientFF, Mollahosseini2019AffectNetAD, Barsoum2016TrainingDN}
    \item
        \textbf{Frame sequence based Deep Learning} \\
        Extract expression or geometric features for each frame in the sequence of a data sample in context of time. Use feature sequences for expression final learning and inference.
        \cite{Liu2014LearningEO, Kahou2015RecurrentNN, Jung2015JointFI}
\end{itemize}

\subsection{Domain Adaptation \& Generalization} \label{subsec:DAG}
According to cross domain scenarios, Domain Adaptation (DA) and Domain Generalization (DG) are addressed sequentially in order of ascending complexity as follows:

\begin{itemize}
    \item 
        \textbf{Domain Adaptation} \\
        The approach of the described methods is mainly segregation of domain specific features to the deepest as possible layers of the network, so that the shallow layers learn transferable task specific knowledge, whereas the ending model structures may be tailored to the target domain by using smaller amount of labeled or pseudo-labeled dataset. The practice of generating samples in the context of lacking data is moderately used, however mostly avoided in newer approaches due to being resource intense and expensive in terms of time. 
        \cite{Long2015learning, Tzeng2015simultaneous, Saenko2010AdaptingVC, Wang2018unsupervised, Goodfellow2015ChallengesIR, Lucey2010ExtendedCK, Ganin2015unsupervised, Bekkouch2019TripletLN, Batanina2019DomainAF}
    \item
        \textbf{Domain Generalization} \\
        Better model generalization entails incorporating more datasets in the learning process. Extending DA approaches towards more domains also involves application of various sophisticated regularization techniques  via, for example, meta-learning or discarding dominant features. Modern design of models is still far from perfect and cannot meet the needs of real use in the industry. Thus, improvements and new approaches are highly demanded.
        \cite{Ghifary2015DomainGF, Saenko2010AdaptingVC, Makhzani2016adversarial, Li2018DomainGW, Li2019EpisodicTF, Balaji2018MetaRegTD, Huang2020self, Li2018LearningTG, Siddiqi2016EvaluatingRP}
\end{itemize}

\subsection{Summary} \label{sec:summary}

To summarize, the choice of a solution strongly depends on the specific problem and the context of its solution. There is always a trade off between many factors to consider when developing and applying systems for Facial Expression Recognition.
The field of the research still remains broad and thus encourages for the search of improvements and new approaches. As observed, the application of existing DA approaches, for example, domain discrimination, to the task of FER is insufficiently studied. To cover the mentioned \textbf{research gap} we pose the research questions:
\begin{enumerate}
    \item What is the improvement of utilizing domain discrimination loss in the learning process for a FER system?
    \item What performance is introduced by separation of domain-invariant and domain specific feature learning neural layers?
\end{enumerate}

\section{Methodology}
The section introduces methods for solving the research problem. Section  \ref{sec:data_preprocessing} studies the operational data and the information preparation techniques, section \ref{sec:architecture} explains the structure of the proposed FER classifier, section \nameref{sec:exp_design} presents the experiment design, whereas other important components of the system are demonstrated in subsequent parts: \nameref{sec:factor}, \nameref{sec:clamping}, \nameref{sec:sampling}.

\subsection{Data preprocessing} \label{sec:data_preprocessing}
Domain Adaptation (DA) task assumes source and target domains to be related to the same task, however, follow different distributions. Our work operates with one source and three target domains modelled by Karolinska Directed Emotional Faces (KDEF) \cite{Lundqvist2015KarolinskaDE}, Extended Cohn-Kanade (CK) \cite{Lucey2010ExtendedCK, CK2000}, Japanese Female Facial Expression Dataset (JAFFE) \cite{Lyons2020CodingFE}, Taiwanese Facial Expression Image Database (TFEID) \cite{TFEID2007} correspondingly.

\begin{figure}[ht]
\centering
\includegraphics[width = \linewidth]{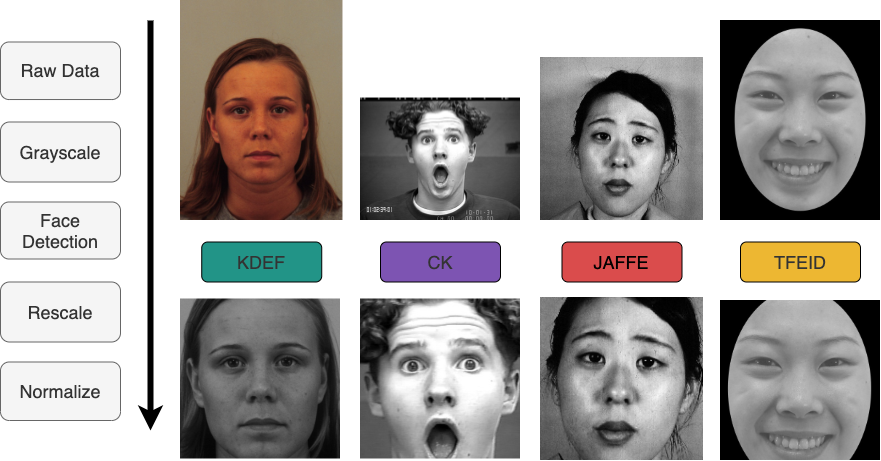}
\caption{Data preprocessing: grayscaling, face detection, rescaling and normalization.}
\label{fig:arch_data}
\end{figure}

Since the data vary significantly (in chromaticity, image size, face positioning, photography method and the degree of post-processing), pre-processing is an important step of the pipeline. As only KDEF is colored, all images are converted to monochrome. In the next steps face detection is applied and photos are rescaled to $224 \times 224$, which is the standard input shape to VGG-19 network used further (Fig. \ref{fig:arch_data})

\subsection{Architecture} \label{sec:architecture}
The architecture of the presented approach represents the expression classifier based on the transfer learning component (feature extracting CNN).
Earlier mentioned VGG19 \cite{Simonyan2015Deep} described in \citeyear{Simonyan2015deep} showed state-of-the-art results in ImageNet \cite{deng2009imagenet} competition in 2014 on object localization and classification. 

Several linear layers follow VGG19 component to adapt and compact features to a lower-dimensional task space. CNNs tend to extract different levels of visual information depending on the deepness of the layer. Hence, features coming from each of the max pooling and the first linear layers of the original VGG are densed to $100$ component vectors each and concatenated to form a single dimensional 600 component vector. Fully connected linear layers are followed by leaky ReLU activation function. Features most significant for facial emotions are expected to be extracted to $600$ dimensional latent space by the feature extractor $F$.

Label predictor compacts the extracted features to 7 neurons representing all considered facial emotions. The last layer utilizes Log Softmax instead of activation to achieve the predicted distribution logits. Negative Log Likelihood ($NLL$) function is used for calculating classification loss and updating the network parameters by the back-propagation of the computed gradients. Classifier conditional distribution of logits $\hat{EXP}|x$, label prediction $\hat{exp}$ and training minimization objective $L_{CLF}$ may be represented as follows:

\begin{equation}
\begin{aligned}
    \widehat{EXP}|x &\sim CLF(F(x))         \\
    \widehat{exp}   &=    argmax[\widehat{EXP}|x] \\
    L_{CLF}         &=    NLL(\widehat{EXP}|x, EXP)
\end{aligned}
\end{equation}

Domain Classifier (DC) is designed to enforce feature extractor to generate domain-invariant features. It is similar to previously mentioned label predictor, however, to achieve the expected result, the domain discrimination loss back-propagated from DC is passed through the Gradient Reversal (GradRev) layer, where it is reversed in direction (multiplied by $-1$). GradRev applies absolutely no action during the forward pass:

\begin{equation}
\begin{aligned}
    GradRev(f) &= f               \\
    \nabla_{f}{GradRev(f)}  &= -1
\end{aligned}
\end{equation}

Consequently, optimization of DC makes features domain-invariant. Respective domain label conditional distribution $\hat{DMN}|x$, domain prediction $\hat{dmn}$ and training minimization objective $L_{DMN}$ may be represented as follows:

\begin{equation}
\begin{aligned}
    \widehat{DMN}|x &\sim DMN(F(x))                \\
    \widehat{dmn}   &=    argmax[\widehat{DMN}|x]  \\
    L_{DMN}         &=    NLL(\widehat{DMN}|x, DMN)
\end{aligned}
\end{equation}

\begin{figure}[ht]
\centering
\includegraphics[width = 0.8\linewidth]{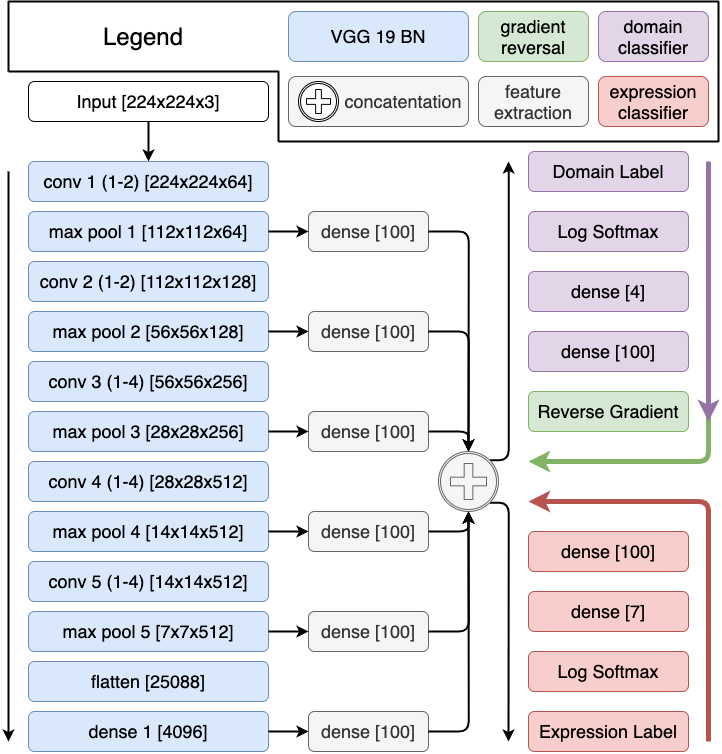}
\caption{Facial Expression Recognition model: detailed architecture. Colored arrows represent the loss back-propagation direction, whereas black ones correspond to the forward pass.}
\label{fig:arch_da_detailed}
\end{figure}

Figure \ref{fig:arch_da_detailed} gives a more detailed representation of the whole system altogether with domain discriminator plugged in.

\subsection{Unsupervised Domain Adaptation} \label{sec:exp_design}
The section describes the design of experiments. The presented domain adaptation model is naturally suitable for the unsupervised training setup. 

Regardless of the use of the proposed domain adaptation technique, the model is granted complete access to the labels of only the source domain and the data origin identifier (domain label). Hence, the experiments provide fair comparison of both settings.

For the unsupervised domain adaptation setup, additional stimulus in the form of domain discrimination loss is present, hence, the domain classifier is activated at each training step. Classification loss is calculated only for the samples from the source domain. In such case, both losses are combined using $\lambda$ hyper-parameter as follows:

\begin{equation}
\begin{aligned}
    L = L_{CLF} + \lambda * L_{DMN} 
\end{aligned}
\end{equation}

Therefore, expected outcomes include:
\begin{enumerate}
    \item feature extractor generates domain-indistinguishable features
    \item trained classifier performs better on target domains
\end{enumerate}

\subsection{Factor function} \label{sec:factor}
In order to reduce the noise coming from domain discrimination loss in the beginning of the training and allow increasing the impact of the domain discriminator gradually, the previously mentioned parameter $\lambda$ is controlled by factor function of the following form:

\begin{equation}
\begin{aligned}
    \lambda(n) = \frac{2}{1 + \exp{(-\frac{\alpha * n}{N}})}-1
\end{aligned}
\end{equation}

where $\frac{n}{N}$ represents the fraction of the steps passed to the total number of steps. Such functional dependency gradually increase $\lambda$ from $0$ to $1$.

\subsection{Scheduled clamping} \label{sec:clamping}

Big emissions of the domain discrimination loss may cause high instability during the training. To decrease the undesired impact and control the growth of the domain loss, we introduce an auxiliary technique called clamping. values exceeding the pre-determined limit convert to the maximal borderline by clamping operation for each term before averaging the loss across mini-batches. The maximum value is controlled by previously mentioned $factor$ function and $clamp$ parameter as follows:

\begin{equation}
\begin{aligned}
    L_{DMN}(n) = max(L_{DMN}, clamp * \lambda(n))
\end{aligned}
\end{equation}

\subsection{Data sampling mechanism} \label{sec:sampling}
Balanced data sampling mechanism is very important in case of domain adaptation scenario. Our approach suggests updating the model parameters after the calculation of the loss for all four datasets at the same time step. For this purpose, mini-batches from all datasets are combined into an aggregated batch before passing in the training step. Such design, notably, increases the training volume because some datasets, being shorter than others, are exploited several times to maintain the inter-domain balance during training. 


\section{Evaluation and Discussion} \label{sec:eval}
The section describes the conducted experiments, presents and discusses the corresponding results. The observations include baseline classifier selection and performance verification on several datasets, fine-tuning of the pre-trained model to target domains and the application of the unsupervised domain adaptation approach. Each of the delivered series of experiments is discussed in detail to explain the obtained results and build the conclusions of the performed work. The paper finishes with a concluding section that summarizes the key findings \nameref{sec:key_findings}.

\subsection{Baseline emotion classifier selection}

Starting experiments build a baseline facial expression classifier. Several datasets, as well as their combinations, are used to validate the performance. Moreover, the model, pre-trained on one of the domains chosen as the source domain (KDEF), is fine-tuned to several source domains (CK, JAFFE, TFEID) in order to investigate its initial ability for domain adaptation.

The experiments train model for \textbf{$100$ epochs}, with a \textbf{batch size} of $40$ images and a \textbf{learning rate} of $1*10^{-4}$. The joint training data source is formed with the use of datasets concatenation and randomized shuffling. Four combinations of domains are chosen: sequentially expanding single KDEF by one dataset at a time up to a set containing all data origins. The data split into $80\%$ training and $20\%$ validation sets.

\begin{table}[ht]
\centering
\begin{tabular}{V{3}c|c|c|c|cV{3}}
\hlineB{3}
       & \multicolumn{4}{cV{3}}{Accuracy} \\ \hline
Domain & KDEF    & \begin{tabular}[c]{@{}c@{}}KDEF\\ CK\end{tabular} & \begin{tabular}[c]{@{}c@{}}KDEF\\ CK\\ JAFFE\end{tabular} & \begin{tabular}[c]{@{}c@{}}KDEF\\ CK\\ JAFFE\\ TFEID\end{tabular} \\ \hline
KDEF   & 87.75\% & \textbf{89.28\%} & 86.22\% & \textbf{89.28\%} \\ \hline
CK     & 48.38\% & \textbf{98.92\%} &\textbf{ 98.92\%} & \textbf{98.92\%} \\ \hline
JAFFE  & 48.83\% & 37.20\% & 81.39\% & \textbf{83.72\%} \\ \hline
TFEID  & 86.95\% & 76.08\% & 73.91\% & \textbf{97.82\%} \\ \hlineB{3}
       & \multicolumn{4}{cV{3}}{Loss} \\ \hline
Domain & KDEF          & \begin{tabular}[c]{@{}c@{}}KDEF\\ CK\end{tabular} & \begin{tabular}[c]{@{}c@{}}KDEF\\ CK\\ JAFFE\end{tabular} & \begin{tabular}[c]{@{}c@{}}KDEF\\ CK\\ JAFFE\\ TFEID\end{tabular} \\ \hline
KDEF   & \textbf{0.55} & 0.59 & 0.59 & 0.68 \\ \hline
CK     & 5.62 & \textbf{0.01} & 0.02 & 0.06 \\ \hline
JAFFE  & 3.16 & 3.86 & 0.79 & \textbf{0.74} \\ \hline
TFEID  & 0.36 & 1.08 & 1.10 & \textbf{0.05} \\ \hlineB{3}
\end{tabular}
\caption{Validation \textbf{accuracy} and \textbf{loss} of the baseline classifier trained on KDEF and three joint domains (columns).}
\label{tab:bc_acc_loss_joint}
\end{table}

Reasonable to assume that the model improves the results only for the new dataset added to the joint training set at a time. However, according to the Table \ref{tab:bc_acc_loss_joint}, one can perceive that the accuracy for KDEF slightly increases ($+1.53\%$) with the extension by CK. Another pair of datasets attend the same idea: the accuracy for JAFFE also increases by $2.32\%$ with the addition of TFEID. Approximately similar results are reflected by the loss indicators represented in Table \ref{tab:bc_acc_loss_joint} as well. It is known that KDEF and CK represent European, while the other two contain Asian faces. Thus, the improvement observed on earlier seen data due to the new data indicates the degree of their similarity.

Based on the results of fine-tuning the model previously trained on source domain KDEF to the target domains, one notes that the retrained classifier loses its original experiences. Consequently, even though the model was previously pre-trained on KDEF, retraining, for example, to CK, causes the accuracy drop from $87.75\%$ to $40.81\%$ on the KDEF, but rises from $48.38\%$ to $98.92\%$ on the target CK dataset (Table \ref{tab:bc_acc_loss_ft}).

\begin{table}[ht]
\centering
\begin{tabular}{V{3}c|c|c|c|cV{3}}
\hlineB{3}
       & \multicolumn{4}{cV{3}}{Accuracy}                                          \\ \hline
Domain & KDEF             & CK               & JAFFE            & TFEID            \\ \hline
KDEF   & \textbf{87.75\%} & 40.81\%          & 75.00\%          & 59.69\%          \\ \hline
CK     & 48.38\%          & \textbf{98.92\%} & 56.20\%          & 32.87\%          \\ \hline
JAFFE  & 48.83\%          & 27.90\%          & \textbf{86.04\%} & 18.60\%          \\ \hline
TFEID  & 86.95\%          & 52.17\%          & 58.69\%          & \textbf{97.82\%} \\ \hlineB{3}
       & \multicolumn{4}{cV{3}}{Loss}                                  \\ \hline
Domain & KDEF          & CK            & JAFFE         & TFEID         \\ \hline
KDEF   & \textbf{0.55} & 7.43          & 2.74          & 4.82          \\ \hline
CK     & 5.62          & \textbf{0.02} & 4.98          & 6.77          \\ \hline
JAFFE  & 3.16          & 6.90          & \textbf{0.95} & 13.57         \\ \hline
TFEID  & 0.36          & 4.24          & 3.50          & \textbf{0.05} \\ \hlineB{3}
\end{tabular}
\caption{Validation \textbf{accuracy} and \textbf{loss} of the baseline classifier fine-tuned to the target domains.}
\label{tab:bc_acc_loss_ft}
\end{table}

\begin{figure}[ht]
    \centering
    \subfloat[training on KDEF]{\includegraphics[width=0.49\linewidth]{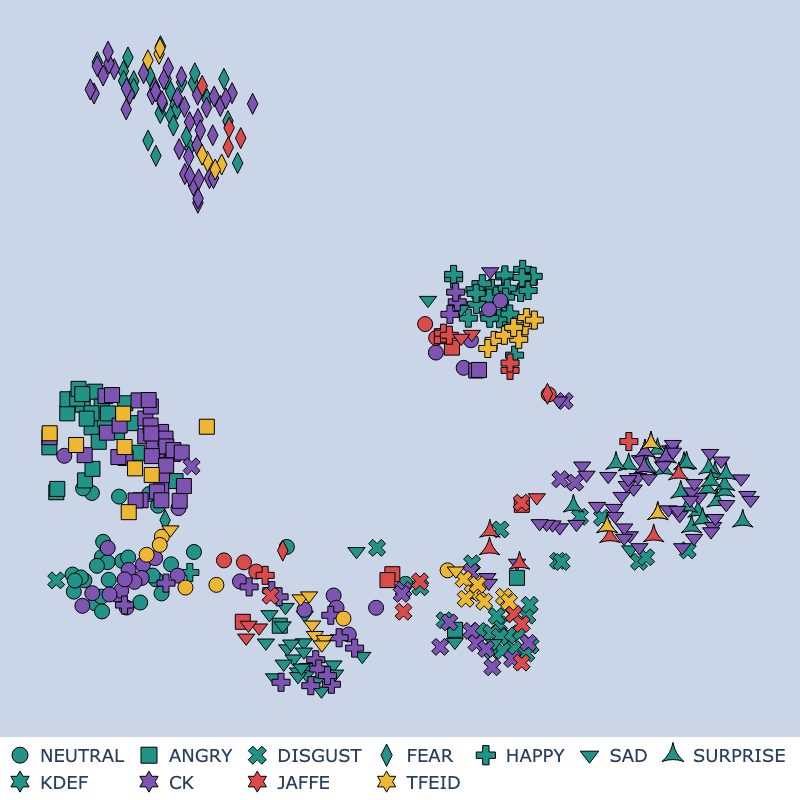}}
    \hspace{0.001\linewidth}
    \subfloat[training on all domains]{\includegraphics[width=0.49\linewidth]{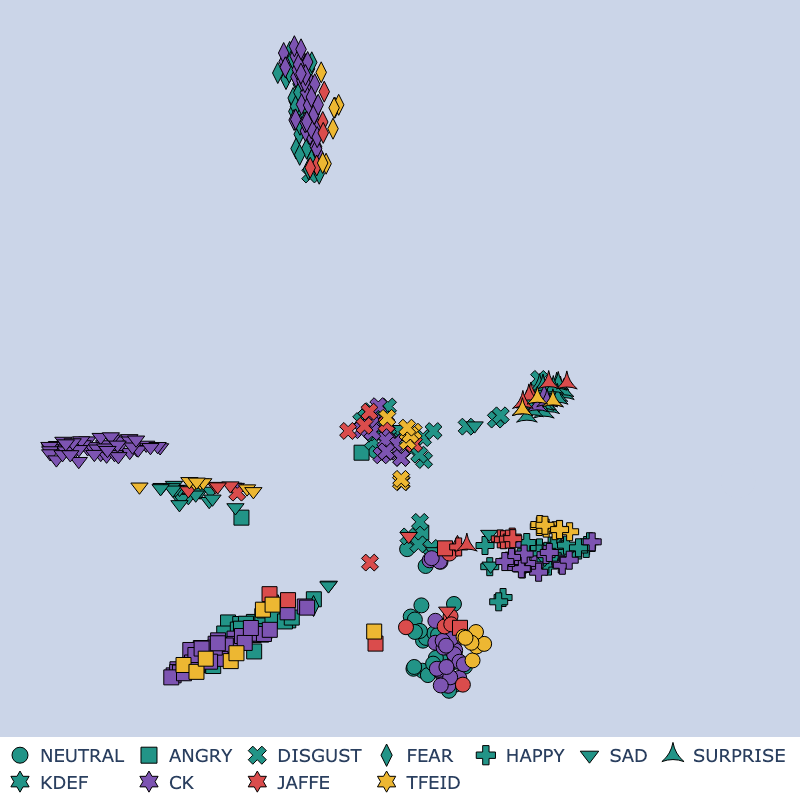}}
    \hspace{0.001\linewidth}
    \subfloat[fine-tuned to CK]{\includegraphics[width=0.49\linewidth]{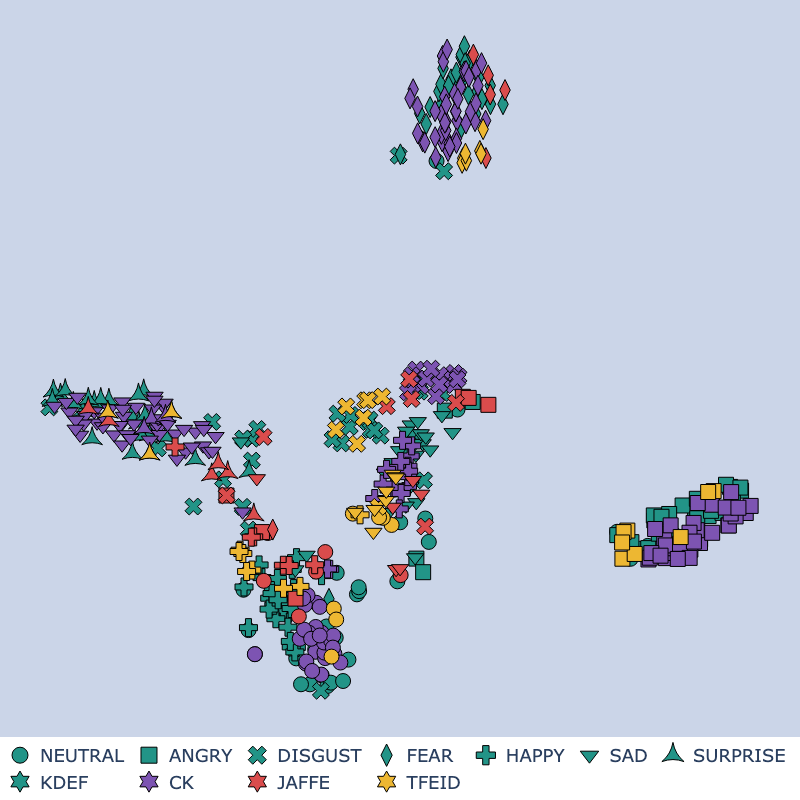}}
    \caption{Latent space visualization of the baseline classifier features.}
    \label{fig:bc_latent}
\end{figure}

T-SNE visualization of the latent space representation of the features displayed in Figure \ref{fig:bc_latent} confirm the results: the model quite expectedly may lose its original skills in favor of a new domain in case they are not continued to be used.

However, the separation of some clusters related to same emotion into groups is also noticeable. For example, $Sad$ emotion samples from CK form a separate group, which is albeit close to the corresponding expression instances from other origins. This observation shows that despite learning to recognize emotions, the model retains the information about the data origin, which makes feature generation less domain-invariant. 

\subsection{Discrimination Loss integration}
The section describes experiments with the use of discrimination loss. 

The hyper-parameters are the following: $100$ epochs; $clamp$ varies in values of $3000$, $5000$, $7000$ and $10000$; $\alpha=10$.

\begin{figure}[ht]
\centering
\includegraphics[width=0.9\linewidth]{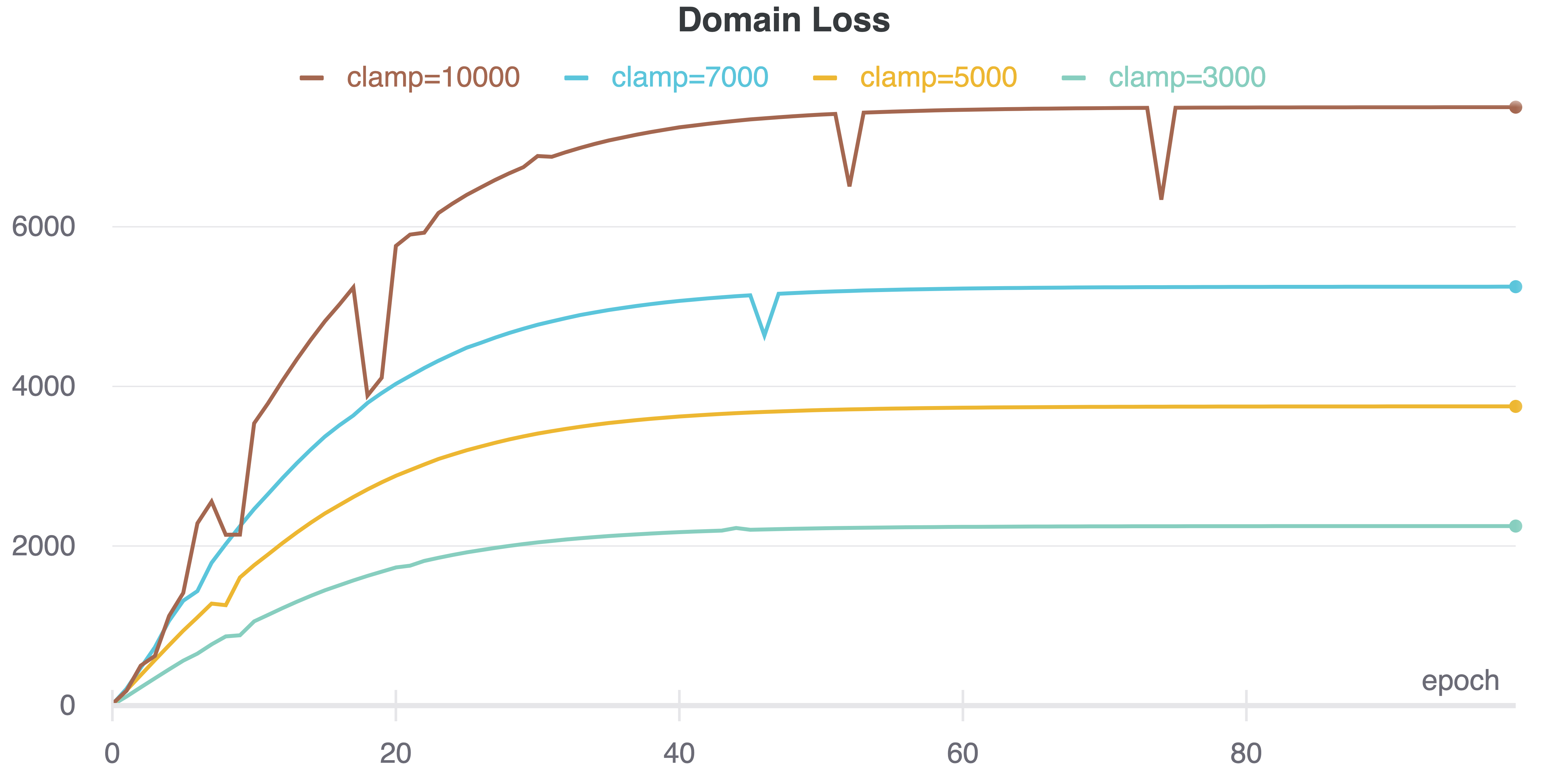}
\caption{Domain discrimination loss: $clamp$ varies in values: $3000$, $5000$, $7000$, $10000$.}
\label{fig:da_final_dmn}
\end{figure}

Figure \ref{fig:da_final_dmn} demonstrates the behavior of the domain discrimination loss, which grows according to the law of the factor function in all cases. It may be noted that fluctuations decrease as the training proceeds and the losses eventually stabilize.

\begin{figure}[ht]
\centering
\includegraphics[width=0.8\linewidth]{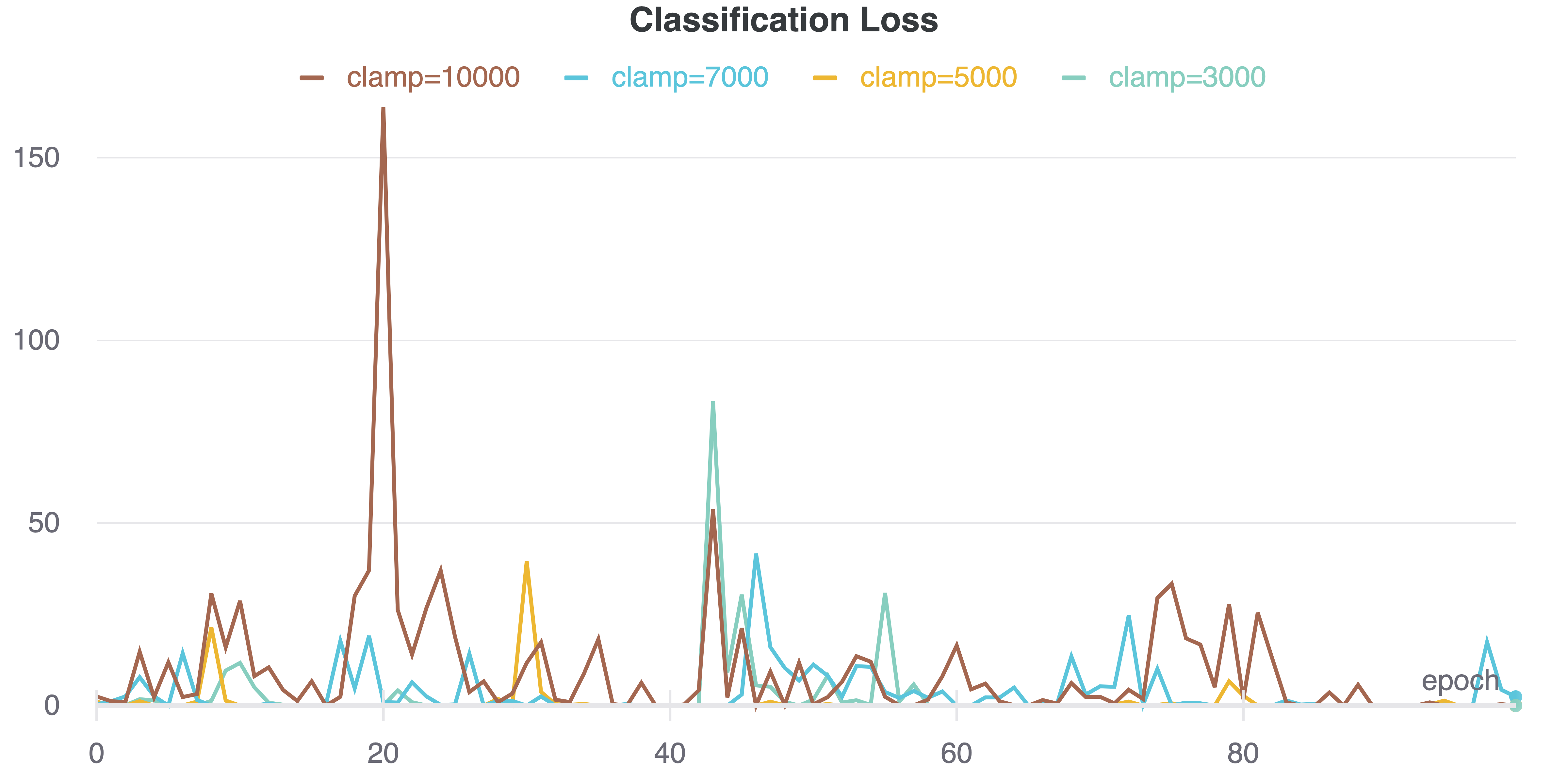}
\caption{Emotion classification loss ($clamp$: $3000$, $5000$, $7000$, $10000$).}
\label{fig:da_final_clf}
\end{figure}

The same situation is reflected in Figure \ref{fig:da_final_clf}: classification loss maintains the downtrend and reaches its minimum values. Thus, the evidence demonstrates that the model converges to a point of stability.

\begin{figure}[ht]
\centering
\subfloat[$clamp=5000$]{\includegraphics[width=0.49\linewidth]{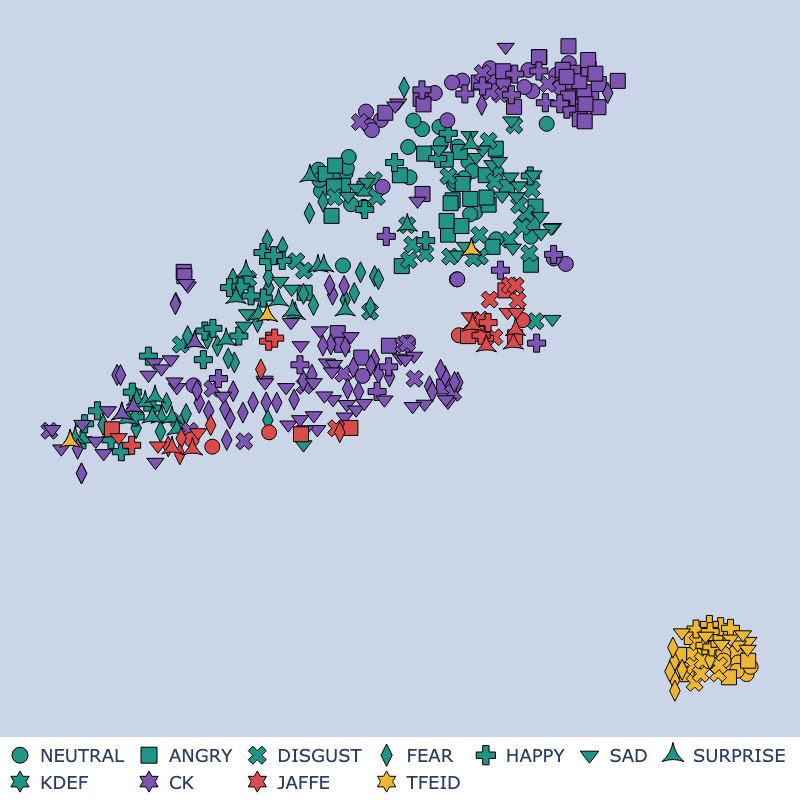}}
\hspace{0.001\linewidth}
\subfloat[$clamp=10000$]{\includegraphics[width=0.49\linewidth]{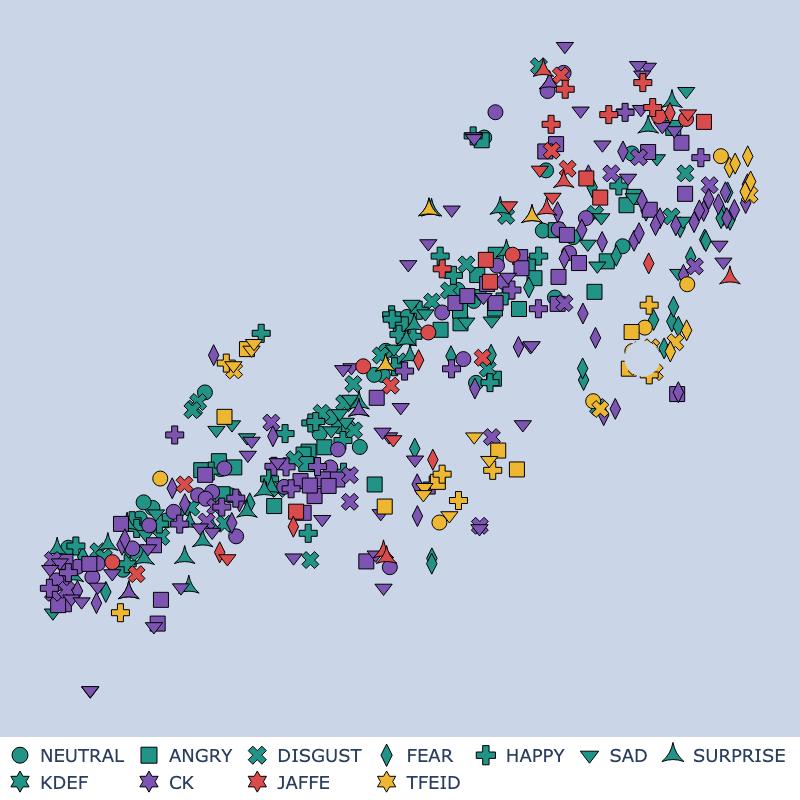}}
\caption{Latent space visualization of features comparison: $clamp$ varies in values: $3000$, $5000$, $7000$, $10000$.}
\label{fig:da_final_latent}
\end{figure}

Figure \ref{fig:da_final_latent} visualizes the latent space of the features obtained in the experiments.

According to the results presented in the Table \ref{tab:da_final}, the best indicators for the accuracy of emotion classification are achieved with $clamp=5000$ for all datasets. As for the latent space representation (\ref{fig:da_final_latent}), the best indistinguishability of data sources is naturally observed at higher values of the parameter $clamp=10000$.

\begin{table}[ht]
\centering
\begin{tabular}{V{3}c|c|c|cV{3}}
\hlineB{3}
       & \multicolumn{3}{cV{3}}{Accuracy}                                   \\ \hline
Domain & $clamp=3000$     & $clamp=5000$     & $clamp=10000$  \\ \hline
KDEF   & 66.32\%          & \textbf{79.08\%} & 72.44\%            \\ \hline
CK     & 51.61\%          & \textbf{52.68\%} & 44.62\%            \\ \hline
JAFFE  & \textbf{25.58\%} & \textbf{25.58\%} & \textbf{25.58\%}   \\ \hline
TFEID  & 28.26\%          & \textbf{32.60\%} & 19.56\%            \\ \hlineB{3}
       & \multicolumn{3}{cV{3}}{Loss}                                       \\ \hline
Domain & $clamp=3000$     & $clamp=5000$ & $clamp=10000$  \\ \hline
KDEF   & \textbf{1.45}  & 2.17           & 4.75                    \\ \hline
CK     & \textbf{6.61}  & 24.72          & 33.22                   \\ \hline
JAFFE  & \textbf{10.29} & 24.57          & 42.88                   \\ \hline
TFEID  & 24.33          & \textbf{14.96} & 30.75                   \\ \hlineB{3}
\end{tabular}
\caption{Validation classification results ($clamp$: $3000$, $5000$, $10000$).}
\label{tab:da_final}
\end{table}

Table \ref{tab:final_comp} compares the validation accuracy provided by the baseline model trained on labeled KDEF and two other candidates utilizing the unlabeled data from other domains through the application of the unsupervised domain adaptation.

\begin{table}[ht]
\centering
\begin{tabular}{V{3}c|c|c|cV{3}}
\hlineB{3}
       & \multicolumn{3}{cV{3}}{Accuracy}                         \\ \hline
Domain & baseline         & $clamp=5000$     & $clamp=10000$      \\ \hline
KDEF   & \textbf{87.75\%} & 79.08\%          & 72.44\%            \\ \hline
CK     & 48.38\%          & \textbf{52.68\%} & 44.62\%            \\ \hline
JAFFE  & \textbf{48.83\%} & 25.58\%          & 25.58\%            \\ \hline
TFEID  & \textbf{86.95\%} & 32.60\%          & 19.56\%            \\ \hlineB{3}
       & \multicolumn{3}{cV{3}}{Loss}                             \\ \hline
Domain & baseline         & $clamp=5000$     & $clamp=10000$      \\ \hline
KDEF   & \textbf{0.55}    & 2.17             & 4.75               \\ \hline
CK     & \textbf{5.62}    & 24.72            & 33.22              \\ \hline
JAFFE  & \textbf{3.16}    & 24.57            & 42.88              \\ \hline
TFEID  & \textbf{0.36}    & 14.96            & 30.75              \\ \hlineB{3}
\end{tabular}
\caption{Comparison of validation results for the baseline model (option A) and domain adapted models using clamp values $5000$ or $10000$.}
\label{tab:final_comp}
\end{table}

The comparison results show that although the model converged towards the point of stability, the approach itself provides insufficient improvements over the baseline model. Despite that fact, \textbf{the key contribution} of the research lies in the \textbf{observed patterns} identified during the whole work. \textbf{Key findings} of the experiments are summarized in the next section.

\section{Conclusion} \label{sec:key_findings}
The section summarizes the key findings identified by the conducted work. 

\subsection{New Architecture for FER}
One of the most important results is the discovery of a new architecture for the task of Facial Expression Recognition. The model referred to in the work as the baseline is an implementation of the popular DL pattern named Transfer Learning. The design is fully end-to-end and demonstrates high validation results in terms of accuracy and loss. The model shows positive dynamics in the source-to-source learning setup and achieves high validation results on several of the popular datasets (KDEF, CK, JAFFE, TFEID). The architecture has shown flexibility in joint dataset training and finetuning environments. Latent space representations confirm the formation of distant clusters and the high availability of the framework to learn facial expressions by distancing the corresponding features. In general, the conducted experiments demonstrate an affirmative experience of using the transfer learning developments for the FER task. The outcome of this part serves as the basis for testing the operation of various optimizations and regularization mechanisms for the improvement of the learning properties. The Domain Adaptation approach attended in the research may be viewed as one of those expansions of the baseline model.

\subsection{Sampling mechanism}
Balancing between domains and aggregated scoring is critical to learning in a domain adaptation context. Our approach, for instance, implies the growth of domain discrimination loss, which is necessary to be balanced. The sampling mechanism chosen in this work assumes the loss averaging over all mini-batches before back-propagation. Mini-batches, in turn, are added to one data packet supplied at one training step to the learning process. Thus, ensuring domain-balanced data is crucial for the learning consistency in a DA setup.

\subsection{Features}

One of the useful observations is that emotion classification features are strongly inter-connected with domain characteristics. According to the conducted studies, any sharp fluctuations in domain discrimination loss negatively affect the classification loss. Therefore, it is highly important to control the smoothness of the domain loss ascending. Moreover, the strong coherence between domain related and emotion related characteristics entails the prediction quality decrease due to forcing the model generating domain-invariant features. Such information loss involves over-fitting to the source domain as the training loss approaches zero but the validation results worsen. One of the useful observations is that while the target domains become less distinguishable and move their data-points towards the original source in latent space representations, the overall features quality decreases.

\subsection{Scheduling function}
The conducted work persistently uses scheduling of several quantities. It is important to ensure the noise coming from the domain discriminator impacts the system less in the beginning of the training process, so the back-propagated gradients are multiplied to the proportion growing from zero to one as the learning progresses. Later, clamping of the domain discrimination loss utilized similar proportioning for providing stability of the training. For the both cases the $factor$, discussed in sub-section \ref{sec:factor} is successfully used as a nonlinear increasing function. Smooth scheduling of important quantities allow for building more stable and consistent systems.

\subsection{Future plans}
As discussed before, despite under-performing results of application of the domain adaptation approach, useful observations allow to build several ideas to explain the behavior and suggest ways for further improvement of the model. 

Decreasing complexity of emotion classifier may reduce over-fitting to the source domain that was examined in last experiments. One of the possible reasons why the feature extractor part remains ineffective at disentangling domain-invariant features could be insufficient capacity of the model. Hence, stacking more layers is a common way to solve problems, however is not preferred considering the initial interest in efficiency. Other methods to solve imperfections consider application of the domain discrimination loss to deeper structures of the feature extractor or the introduction of additional stimuli and regularization techniques.

To conclude, although the conducted experiments cannot provide the world with a new competitive solution, they discovered useful observations. The further analysis of the findings allows to determine the directions for further research and improvements.

\printbibliography[heading=bibintoc,title={References}]

@inproceedings{Lucey2010ExtendedCK,
  author={P. {Lucey} and J. F. {Cohn} and T. {Kanade} and J. {Saragih} and Z. {Ambadar} and I. {Matthews}},
  booktitle={2010 IEEE Computer Society Conference on Computer Vision and Pattern Recognition - Workshops}, 
  title={The Extended Cohn-Kanade Dataset (CK+): A complete dataset for action unit and emotion-specified expression}, 
  year={2010},
  volume={},
  number={},
  pages={94-101},
  doi={10.1109/CVPRW.2010.5543262}}

@article{Susskind2010toronto,
  title={The toronto face database},
  author={Susskind, Josh M and Anderson, Adam K and Hinton, Geoffrey E},
  journal={Department of Computer Science, University of Toronto, Toronto, ON, Canada, Tech. Rep},
  volume={3},
  year={2010}
}

@article{Mavadati2013DISFA,
    author = {Mavadati, Seyedmohammad and Mahoor, Mohammad and Bartlett, Kevin and Trinh, Philip and Cohn, Jeffrey},
    year = {2013},
    month = {04},
    pages = {151-160},
    title = {DISFA: A spontaneous facial action intensity database},
    volume = {4},
    journal = {Affective Computing, IEEE Transactions on},
    doi = {10.1109/T-AFFC.2013.4}
}

@inproceedings{Lundqvist2015KarolinskaDE,
  title={Karolinska Directed Emotional Faces},
  author={D. Lundqvist and A. Flykt and A. {\"O}hman},
  year={2015}
}

@inproceedings{Pantic2005wdffe,
    author = {M. Pantic and M. F. Valstar and R. Rademaker and L. Maat},
    pages = {317--321},
    address = {Amsterdam, The Netherlands},
    booktitle = {Proceedings of IEEE Int'l Conf. Multimedia and Expo (ICME'05)},
    month = {July},
    title = {Web-based database for facial expression analysis},
    year = {2005},
}

@INPROCEEDINGS{Gross2008MultiPIE,
  author={R. {Gross} and I. {Matthews} and J. {Cohn} and T. {Kanade} and S. {Baker}},
  booktitle={2008 8th IEEE International Conference on Automatic Face   Gesture Recognition}, 
  title={Multi-PIE}, 
  year={2008},
  volume={},
  number={},
  pages={1-8},
  doi={10.1109/AFGR.2008.4813399}
}

@article{Goodfellow2015ChallengesIR,
  title={Challenges in representation learning: A report on three machine learning contests},
  author={Ian J. Goodfellow and D. Erhan and Pierre Luc Carrier and Aaron C. Courville and M. Mirza and Benjamin Hamner and William Cukierski and Y. Tang and D. Thaler and Dong-Hyun Lee and Yingbo Zhou and Chetan Ramaiah and Fangxiang Feng and Ruifan Li and X. Wang and Dimitris Athanasakis and J. Shawe-Taylor and Maxim Milakov and John Park and Radu Tudor Ionescu and M. Popescu and C. Grozea and J. Bergstra and Jingjing Xie and Lukasz Romaszko and Bing Xu and Chuang Zhang and Yoshua Bengio},
  journal={Neural networks : the official journal of the International Neural Network Society},
  year={2015},
  volume={64},
  pages={
          59-63
        }
}

@inproceedings{Li2013OuluCasia,
author = {Li, Stan and Yi, Dong and Lei, Zhen and Liao, Shengcai},
year = {2013},
month = {06},
pages = {348-353},
title = {The CASIA NIR-VIS 2.0 face database},
journal = {IEEE Computer Society Conference on Computer Vision and Pattern Recognition Workshops},
doi = {10.1109/CVPRW.2013.59}
}

@article{Mollahosseini2019AffectNetAD,
  title={AffectNet: A Database for Facial Expression, Valence, and Arousal Computing in the Wild},
  author={A. Mollahosseini and B. Hasani and M. Mahoor},
  journal={IEEE Transactions on Affective Computing},
  year={2019},
  volume={10},
  pages={18-31}
}

@article{Barsoum2016TrainingDN,
  title={Training deep networks for facial expression recognition with crowd-sourced label distribution},
  author={E. Barsoum and C. Zhang and C. Canton-Ferrer and Zhengyou Zhang},
  journal={Proceedings of the 18th ACM International Conference on Multimodal Interaction},
  year={2016}
}

@article{Lyons2020CodingFE,
  title={Coding Facial Expressions with Gabor Wavelets (IVC Special Issue)},
  author={M. Lyons and M. Kamachi and J. Gyoba},
  journal={ArXiv},
  year={2020},
  volume={abs/2009.05938}
}

@article{TFEID2007,
  title={Taiwanese  facial  expression image  database},
  author={L.F.  Chen  and  Y.S.  Yen},
  address = {Taipei, Taiwan},
  journal={Brain  Mapping  Laboratory,  Institute  of Brain  Science,  National  Yang-Ming  University},
  year={2007},
}

@inproceedings{Saenko2010AdaptingVC,
  title={Adapting Visual Category Models to New Domains},
  author={Kate Saenko and B. Kulis and M. Fritz and Trevor Darrell},
  booktitle={ECCV},
  year={2010}
}

@inproceedings{Li2018LearningTG,
  title={Learning to Generalize: Meta-Learning for Domain Generalization},
  author={Da Li and Yongxin Yang and Yi-Zhe Song and Timothy M. Hospedales},
  booktitle={AAAI},
  year={2018}
}

@INPROCEEDINGS{CK2000,
  author={Kanade, T. and Cohn, J.F. and Yingli Tian},
  booktitle={Proceedings Fourth IEEE International Conference on Automatic Face and Gesture Recognition (Cat. No. PR00580)}, 
  title={Comprehensive database for facial expression analysis}, 
  year={2000},
  volume={},
  number={},
  pages={46-53},
  doi={10.1109/AFGR.2000.840611}}

@incollection{Schapire2013explaining,
  title={Explaining adaboost},
  author={Schapire, Robert E},
  booktitle={Empirical inference},
  pages={37--52},
  year={2013},
  publisher={Springer}
}

@Article{Liaw2002RandomForest,
    title = {Classification and Regression by randomForest},
    author = {Andy Liaw and Matthew Wiener},
    journal = {R News},
    year = {2002},
    volume = {2},
    number = {3},
    pages = {18-22},
    url = {https://CRAN.R-project.org/doc/Rnews/},
}

@article{Cortes1995Support,
  title={Support-vector networks},
  author={Cortes, Corinna and Vapnik, Vladimir},
  journal={Machine learning},
  volume={20},
  number={3},
  pages={273--297},
  year={1995},
  publisher={Springer}
}

@inproceedings{Chen2014FacialER,
  title={Facial Expression Recognition Based on Facial Components Detection and HOG Features},
  author={Junkai Chen and Zenghai Chen and Z. Chi and Hong Fu},
  year={2014}
}

@article{Shan2009FacialER,
  title={Facial expression recognition based on Local Binary Patterns: A comprehensive study},
  author={Caifeng Shan and S. Gong and P. McOwan},
  journal={Image Vis. Comput.},
  year={2009},
  volume={27},
  pages={803-816}
}

@article{Bartlett2005RecognizingFE,
  title={Recognizing facial expression: machine learning and application to spontaneous behavior},
  author={M. Bartlett and Gwen Littlewort and M. Frank and C. Lainscsek and Ian R. Fasel and J. Movellan},
  journal={2005 IEEE Computer Society Conference on Computer Vision and Pattern Recognition (CVPR'05)},
  year={2005},
  volume={2},
  pages={568-573 vol. 2}
}

@article{Whitehill2006HaarFF,
  title={Haar features for FACS AU recognition},
  author={Jacob Whitehill and C. Omlin},
  journal={7th International Conference on Automatic Face and Gesture Recognition (FGR06)},
  year={2006},
  pages={5 pp.-101}
}

@article{Suk2014RealTimeMF,
  title={Real-Time Mobile Facial Expression Recognition System -- A Case Study},
  author={M. Suk and B. Prabhakaran},
  journal={2014 IEEE Conference on Computer Vision and Pattern Recognition Workshops},
  year={2014},
  pages={132-137}
}

@article{Ghimire2013GeometricFF,
  title={Geometric Feature-Based Facial Expression Recognition in Image Sequences Using Multi-Class AdaBoost and Support Vector Machines},
  author={D. Ghimire and J. Lee},
  journal={Sensors (Basel, Switzerland)},
  year={2013},
  volume={13},
  pages={7714 - 7734}
}

@article{Siddiqi2015HumanFE,
  title={Human Facial Expression Recognition Using Stepwise Linear Discriminant Analysis and Hidden Conditional Random Fields},
  author={M. Siddiqi and R. Ali and A. Khan and Young-Tack Park and S. Lee},
  journal={IEEE Transactions on Image Processing},
  year={2015},
  volume={24},
  pages={1386-1398}
}

@article{Khan2013FrameworkFR,
  title={Framework for reliable, real-time facial expression recognition for low resolution images},
  author={R. A. Khan and A. Meyer and H. Konik and S. Bouakaz},
  journal={Pattern Recognit. Lett.},
  year={2013},
  volume={34},
  pages={1159-1168}
}

@article{Ghimire2016FacialER,
  title={Facial expression recognition based on local region specific features and support vector machines},
  author={D. Ghimire and SungHwan Jeong and J. Lee and S. Park},
  journal={Multimedia Tools and Applications},
  year={2016},
  volume={76},
  pages={7803-7821}
}

@article{BenitezQuiroz2016EmotioNetAA,
  title={EmotioNet: An Accurate, Real-Time Algorithm for the Automatic Annotation of a Million Facial Expressions in the Wild},
  author={C. F. Benitez-Quiroz and R. Srinivasan and A. Mart{\'i}nez},
  journal={2016 IEEE Conference on Computer Vision and Pattern Recognition (CVPR)},
  year={2016},
  pages={5562-5570}
}

@INPROCEEDINGS{Torre2015IntraFace,
  author={F. {De la Torre} and W. {Chu} and X. {Xiong} and F. {Vicente} and X. {Ding} and J. {Cohn}},
  booktitle={2015 11th IEEE International Conference and Workshops on Automatic Face and Gesture Recognition (FG)}, 
  title={IntraFace}, 
  year={2015},
  volume={1},
  number={},
  pages={1-8},
  doi={10.1109/FG.2015.7163082}
}

@article{Khorrami2015DoDN,
  title={Do Deep Neural Networks Learn Facial Action Units When Doing Expression Recognition?},
  author={Pooya Khorrami and T. Paine and T. Huang},
  journal={2015 IEEE International Conference on Computer Vision Workshop (ICCVW)},
  year={2015},
  pages={19-27}
}

@article{Memisevic2015ZerobiasAA,
  title={Zero-bias autoencoders and the benefits of co-adapting features},
  author={R. Memisevic and Kishore Reddy Konda and David Krueger},
  journal={CoRR},
  year={2015},
  volume={abs/1402.3337}
}

@article{Paine2015AnAO,
  title={An Analysis of Unsupervised Pre-training in Light of Recent Advances},
  author={T. Paine and Pooya Khorrami and Wei Han and T. Huang},
  journal={CoRR},
  year={2015},
  volume={abs/1412.6597}
}

@inproceedings{Aneja2016ModelingSC,
  title={Modeling Stylized Character Expressions via Deep Learning},
  author={Deepali Aneja and Alex Colburn and Gary Faigin and L. Shapiro and Barbara Mones},
  booktitle={ACCV},
  year={2016}
}

@article{Mollahosseini2016GoingDI,
  title={Going deeper in facial expression recognition using deep neural networks},
  author={A. Mollahosseini and D. Chan and M. Mahoor},
  journal={2016 IEEE Winter Conference on Applications of Computer Vision (WACV)},
  year={2016},
  pages={1-10}
}

@article{Liu2014FacialER,
  title={Facial Expression Recognition via a Boosted Deep Belief Network},
  author={Ping Liu and Shizhong Han and Zibo Meng and Yan Tong},
  journal={2014 IEEE Conference on Computer Vision and Pattern Recognition},
  year={2014},
  pages={1805-1812}
}

@article{Minaee2019DeepEmotionFE,
  title={Deep-Emotion: Facial Expression Recognition Using Attentional Convolutional Network},
  author={Shervin Minaee and AmirAli Abdolrashidi},
  journal={ArXiv},
  year={2019},
  volume={abs/1902.01019}
}

@article{Li2019OcclusionAF,
  title={Occlusion Aware Facial Expression Recognition Using CNN With Attention Mechanism},
  author={Y. Li and Jiabei Zeng and S. Shan and X. Chen},
  journal={IEEE Transactions on Image Processing},
  year={2019},
  volume={28},
  pages={2439-2450}
}

@article{Cugu2019MicroExpNetAE,
  title={MicroExpNet: An Extremely Small and Fast Model For Expression Recognition From Face Images},
  author={Ilke Çugu and Eren Sener and Emre Akbas},
  journal={2019 Ninth International Conference on Image Processing Theory, Tools and Applications (IPTA)},
  year={2019},
  pages={1-6}
}

@article{Zhao2016PeakPilotedDN,
  title={Peak-Piloted Deep Network for Facial Expression Recognition},
  author={X. Zhao and Xiaodan Liang and Luoqi Liu and Teng Li and Yugang Han and N. Vasconcelos and S. Yan},
  journal={ArXiv},
  year={2016},
  volume={abs/1607.06997}
}

@article{Szegedy2015GoingDW,
  title={Going deeper with convolutions},
  author={Christian Szegedy and W. Liu and Y. Jia and Pierre Sermanet and Scott Reed and Dragomir Anguelov and D. Erhan and V. Vanhoucke and Andrew Rabinovich},
  journal={2015 IEEE Conference on Computer Vision and Pattern Recognition (CVPR)},
  year={2015},
  pages={1-9}
}

@article{Ding2017FaceNet2ExpNetRA,
  title={FaceNet2ExpNet: Regularizing a Deep Face Recognition Net for Expression Recognition},
  author={H. Ding and S. Zhou and R. Chellappa},
  journal={2017 12th IEEE International Conference on Automatic Face \& Gesture Recognition (FG 2017)},
  year={2017},
  pages={118-126}
}

@article{Ba2014DoDN,
  title={Do Deep Nets Really Need to be Deep?},
  author={Jimmy Ba and R. Caruana},
  journal={ArXiv},
  year={2014},
  volume={abs/1312.6184}
}

@inproceedings{Parkhi2015DeepFR,
  title={Deep Face Recognition},
  author={Omkar M. Parkhi and A. Vedaldi and A. Zisserman},
  booktitle={BMVC},
  year={2015}
}

@article{Kim2017DeepGN,
  title={Deep generative-contrastive networks for facial expression recognition},
  author={Y. Kim and ByungIn Yoo and Youngjun Kwak and C. Choi and Junmo Kim},
  journal={ArXiv},
  year={2017},
  volume={abs/1703.07140}
}

@article{Liu2014LearningEO,
  title={Learning Expressionlets on Spatio-temporal Manifold for Dynamic Facial Expression Recognition},
  author={M. Liu and S. Shan and R. Wang and X. Chen},
  journal={2014 IEEE Conference on Computer Vision and Pattern Recognition},
  year={2014},
  pages={1749-1756}
}

@inproceedings{Kahou2015RecurrentNN,
  title={Recurrent Neural Networks for Emotion Recognition in Video},
  author={S. Kahou and Vincent Michalski and Kishore Reddy Konda and R. Memisevic and C. Pal},
  booktitle={ICMI '15},
  year={2015}
}

@article{Jung2015JointFI,
  title={Joint Fine-Tuning in Deep Neural Networks for Facial Expression Recognition},
  author={Heechul Jung and Sihaeng Lee and Junho Yim and Sunjeong Park and Junmo Kim},
  journal={2015 IEEE International Conference on Computer Vision (ICCV)},
  year={2015},
  pages={2983-2991}
}

@inproceedings{Siqueira2020EfficientFF,
  title={Efficient Facial Feature Learning with Wide Ensemble-based Convolutional Neural Networks},
  author={Henrique Siqueira and Sven Magg and Stefan Wermter},
  booktitle={AAAI},
  year={2020}
}

@misc{Bengio2014representation,
      title={Representation Learning: A Review and New Perspectives}, 
      author={Yoshua Bengio and Aaron Courville and Pascal Vincent},
      year={2014},
      eprint={1206.5538},
      archivePrefix={arXiv},
      primaryClass={cs.LG}
}

@article{Liu2018ExploringDF,
  title={Exploring Disentangled Feature Representation Beyond Face Identification},
  author={Y. Liu and Fangyin Wei and J. Shao and Lu Sheng and J. Yan and X. Wang},
  journal={2018 IEEE/CVF Conference on Computer Vision and Pattern Recognition},
  year={2018},
  pages={2080-2089}
}

@misc{Halawa2020learning,
      title={Learning Disentangled Expression Representations from Facial Images}, 
      author={Marah Halawa and Manuel Wöllhaf and Eduardo Vellasques and Urko Sánchez Sanz and Olaf Hellwich},
      year={2020},
      eprint={2008.07001},
      archivePrefix={arXiv},
      primaryClass={cs.CV}
}

@inproceedings{Zhang2018IdentitybasedAT,
  title={Identity-based Adversarial Training of Deep CNNs for Facial Action Unit Recognition},
  author={Zheng Zhang and Shuangfei Zhai and L. Yin},
  booktitle={BMVC},
  year={2018}
}

@article{Tu2019IdenNetIF,
  title={IdenNet: Identity-Aware Facial Action Unit Detection},
  author={Cheng-Hao Tu and Chih-Yuan Yang and J. Hsu},
  journal={2019 14th IEEE International Conference on Automatic Face \& Gesture Recognition (FG 2019)},
  year={2019},
  pages={1-8}
}

@misc{Ali2019facial,
      title={Facial Expression Recognition Using Disentangled Adversarial Learning}, 
      author={Kamran Ali and Charles E. Hughes},
      year={2019},
      eprint={1909.13135},
      archivePrefix={arXiv},
      primaryClass={cs.CV}
}

@misc{Cai2019identityfree,
      title={Identity-Free Facial Expression Recognition using conditional Generative Adversarial Network}, 
      author={Jie Cai and Zibo Meng and Ahmed Shehab Khan and Zhiyuan Li and James O'Reilly and Yan Tong},
      year={2019},
      eprint={1903.08051},
      archivePrefix={arXiv},
      primaryClass={cs.CV}
}

@article{Xu2019PersonindependentFE,
  title={Person-independent facial expression recognition method based on improved Wasserstein generative adversarial networks in combination with identity aware},
  author={Caie Xu and Yang Cui and Yunhui Zhang and P. Gao and Jiayi Xu},
  journal={Multimedia Systems},
  year={2019},
  volume={26},
  pages={53-61}
}

@inproceedings{Hinton1993AutoencodersMD,
  title={Autoencoders, Minimum Description Length and Helmholtz Free Energy},
  author={Geoffrey E. Hinton and R. Zemel},
  booktitle={NIPS},
  year={1993}
}

@inproceedings{Long2015learning,
  title={Learning transferable features with deep adaptation networks},
  author={Long, Mingsheng and Cao, Yue and Wang, Jianmin and Jordan, Michael},
  booktitle={International conference on machine learning},
  pages={97--105},
  year={2015},
  organization={PMLR}
}

@inproceedings{Ganin2015unsupervised,
  title={Unsupervised domain adaptation by backpropagation},
  author={Ganin, Yaroslav and Lempitsky, Victor},
  booktitle={International conference on machine learning},
  pages={1180--1189},
  year={2015},
  organization={PMLR}
}

@inproceedings{Tzeng2015simultaneous,
  title={Simultaneous deep transfer across domains and tasks},
  author={Tzeng, Eric and Hoffman, Judy and Darrell, Trevor and Saenko, Kate},
  booktitle={Proceedings of the IEEE International Conference on Computer Vision},
  pages={4068--4076},
  year={2015}
}

@article{Wang2018unsupervised,
  title={Unsupervised domain adaptation for facial expression recognition using generative adversarial networks},
  author={Wang, Xiaoqing and Wang, Xiangjun and Ni, Yubo},
  journal={Computational intelligence and neuroscience},
  volume={2018},
  year={2018},
  publisher={Hindawi}
}

@article{Ghifary2015DomainGF,
  title={Domain Generalization for Object Recognition with Multi-task Autoencoders},
  author={Muhammad Ghifary and W. Kleijn and M. Zhang and D. Balduzzi},
  journal={2015 IEEE International Conference on Computer Vision (ICCV)},
  year={2015},
  pages={2551-2559}
}

@article{Li2018DomainGW,
  title={Domain Generalization with Adversarial Feature Learning},
  author={Haoliang Li and Sinno Jialin Pan and S. Wang and A. Kot},
  journal={2018 IEEE/CVF Conference on Computer Vision and Pattern Recognition},
  year={2018},
  pages={5400-5409}
}

@misc{Makhzani2016adversarial,
      title={Adversarial Autoencoders}, 
      author={Alireza Makhzani and Jonathon Shlens and Navdeep Jaitly and Ian Goodfellow and Brendan Frey},
      year={2016},
      eprint={1511.05644},
      archivePrefix={arXiv},
      primaryClass={cs.LG}
}

@article{Li2019EpisodicTF,
  title={Episodic Training for Domain Generalization},
  author={Da Li and J. Zhang and Yongxin Yang and Cong Liu and Yi-Zhe Song and Timothy M. Hospedales},
  journal={2019 IEEE/CVF International Conference on Computer Vision (ICCV)},
  year={2019},
  pages={1446-1455}
}

@inproceedings{Balaji2018MetaRegTD,
  title={MetaReg: Towards Domain Generalization using Meta-Regularization},
  author={Y. Balaji and S. Sankaranarayanan and R. Chellappa},
  booktitle={NeurIPS},
  year={2018}
}

@article{Huang2020self,
  title={Self-challenging improves cross-domain generalization},
  author={Huang, Zeyi and Wang, Haohan and Xing, Eric P and Huang, Dong},
  journal={arXiv preprint arXiv:2007.02454},
  year={2020}
}

@misc{Simonyan2015Deep,
      title={Very Deep Convolutional Networks for Large-Scale Image Recognition}, 
      author={Karen Simonyan and Andrew Zisserman},
      year={2015},
      eprint={1409.1556},
      archivePrefix={arXiv},
      primaryClass={cs.CV}
}

@inproceedings{Arjovsky2017WassersteinGA,
  title={Wasserstein Generative Adversarial Networks},
  author={Mart{\'i}n Arjovsky and Soumith Chintala and L. Bottou},
  booktitle={ICML},
  year={2017}
}

@article{Siddiqi2016ANM,
  title={A Novel Maximum Entropy Markov Model for Human Facial Expression Recognition},
  author={M. Siddiqi and Md. Golam Rabiul Alam and C. Hong and A. Khan and H. Choo},
  journal={PLoS ONE},
  year={2016},
  volume={11}
}

@article{Siddiqi2014HumanFE,
  title={Human facial expression recognition using curvelet feature extraction and normalized mutual information feature selection},
  author={M. Siddiqi and R. Ali and M. Idris and A. Khan and E. Kim and M. C. Whang and S. Lee},
  journal={Multimedia Tools and Applications},
  year={2014},
  volume={75},
  pages={935-959}
}

@article{Siddiqi2014DepthCF,
  title={Depth Camera-Based Facial Expression Recognition System Using Multilayer Scheme},
  author={M. Siddiqi and R. Ali and A. Sattar and A. Khan and S. Lee},
  journal={IETE Technical Review},
  year={2014},
  volume={31},
  pages={277 - 286}
}

@article{Siddiqi2014FacialER,
  title={Facial expression recognition using active contour-based face detection, facial movement-based feature extraction, and non-linear feature selection},
  author={M. Siddiqi and R. Ali and A. Khan and E. Kim and G. Kim and S. Lee},
  journal={Multimedia Systems},
  year={2014},
  volume={21},
  pages={541-555}
}

@article{Siddiqi2013HierarchicalRS,
  title={Hierarchical Recognition Scheme for Human Facial Expression Recognition Systems},
  author={M. Siddiqi and S. Lee and Y. Lee and A. Khan and P. T. H. Truc},
  journal={Sensors (Basel, Switzerland)},
  year={2013},
  volume={13},
  pages={16682 - 16713}
}

@article{Bekkouch2019TripletLN,
  title={Triplet Loss Network for Unsupervised Domain Adaptation},
  author={I. E. I. Bekkouch and Y. Youssry and R. Gafarov and A. Khan and A. Khattak},
  journal={Algorithms},
  year={2019},
  volume={12},
  pages={96}
}

@article{Batanina2019DomainAF,
  title={Domain Adaptation for Car Accident Detection in Videos},
  author={Elizaveta Batanina and I. E. I. Bekkouch and A. Khan and A. Khattak and M. Bortnikov},
  journal={2019 Ninth International Conference on Image Processing Theory, Tools and Applications (IPTA)},
  year={2019},
  pages={1-6}
}

@article{Siddiqi2016EvaluatingRP,
  title={Evaluating real-life performance of the state-of-the-art in facial expression recognition using a novel YouTube-based datasets},
  author={M. Siddiqi and Maqbool Ali and M. Eldib and A. Khan and O. Ba{\~n}os and S. Lee and H. Choo},
  journal={Multimedia Tools and Applications},
  year={2016},
  volume={77},
  pages={917-937}
}

@article{Li2018DeepFE,
  title={Deep Facial Expression Recognition: A Survey},
  author={Shan Li and W. Deng},
  journal={ArXiv},
  year={2018},
  volume={abs/1804.08348}
}

@article{Corneanu2016SurveyOR,
  title={Survey on RGB, 3D, Thermal, and Multimodal Approaches for Facial Expression Recognition: History, Trends, and Affect-Related Applications},
  author={C. Corneanu and Marc Oliu and J. Cohn and S. Escalera},
  journal={IEEE Transactions on Pattern Analysis and Machine Intelligence},
  year={2016},
  volume={38},
  pages={1548-1568}
}

@article{Wu2018FacialLD,
  title={Facial Landmark Detection: A Literature Survey},
  author={Y. Wu and Q. Ji},
  journal={International Journal of Computer Vision},
  year={2018},
  volume={127},
  pages={115-142}
}

@inproceedings{deng2009imagenet,
  title={Imagenet: A large-scale hierarchical image database},
  author={Deng, Jia and Dong, Wei and Socher, Richard and Li, Li-Jia and Li, Kai and Fei-Fei, Li},
  booktitle={2009 IEEE conference on computer vision and pattern recognition},
  pages={248--255},
  year={2009},
  organization={Ieee}
}
\end{document}